\documentclass[runningheads]{llncs}
\usepackage{graphicx}
\usepackage{booktabs}
\begin{document}
\title{Exploring Out-of-Distribution Generalization in Text Classifiers Trained on Tobacco-3482 and RVL-CDIP}
\titlerunning{Exploring Out-of-Distribution Generalization in Text Classifiers}
\authorrunning{S. Larson, et al.}
%
\author{Stefan Larson\inst{1} \and
Navtej Singh\inst{1} \and
Saarthak Maheshwari\inst{2}\thanks{Work performed while author was an intern at SkySync.} \and\\
Shanti Stewart\inst{3}$^\star$ \and 
Uma Krishnaswamy\inst{2}$^\star$}
%
%
\institute{SkySync, Ann Arbor, MI, USA\\
\email{\{slarson, nsingh\}@skysync.com} \and
University of California, Berkeley, CA, USA \and 
Oregon State University, Corvallis, OR, USA}
\maketitle              
\begin{abstract}
To be robust enough for widespread adoption, document analysis systems involving machine learning models must be able to respond correctly to inputs that fall outside of the data distribution that was used to generate the data on which the models were trained.
This paper explores the ability of text classifiers trained on standard document classification datasets to generalize to out-of-distribution documents at inference time.
We take the Tobacco-3482 and RVL-CDIP datasets as a starting point and generate new out-of-distribution evaluation datasets in order to analyze the generalization performance of models trained on these standard datasets.
We find that models trained on the smaller Tobacco-3482 dataset perform poorly on our new out-of-distribution data, while text classification models trained on the larger RVL-CDIP exhibit smaller performance drops.

\end{abstract}
\section{Introduction}
Recent years have seen great improvements in the document analysis and recognition field. These advancements have typically stemmed from higher capacity deep learning models and larger training datasets. Two representative datasets in the field of document classification are the Tobacco-3482 \cite{tobacco_3482} and the RVL-CDIP \cite{rvl_cdip} datasets, the latter consisting of over 400,000 training samples across 16 document categories. Indeed, the RVL-CDIP dataset has emerged as the premier benchmark for evaluating document classification algorithms.

However, there are some potential weaknesses with both of these datasets. For one, documents from both datasets come from the same domain: the tobacco industry. Second, the documents in both corpora are from the 1950s to 2002, and are older than---for lack of a better word---contemporary documents. These two potential weaknesses lead us to question whether models trained on RVL-CDIP and Tobacco-3482 have the capacity to generalize to \emph{out-of-distribution} inputs. That is to say, can models trained on RVL-CDIP and Tobacco-3482 perform well on documents that come from different industries than the tobacco industry? and can such models perform well on more recent documents?

We attempt to answer these questions in this paper. We train text classifiers on the RVL-CDIP and Tobacco-3482 datasets and evaluate performance on new test datasets. These test datasets are drawn from ``distributions" that are quite different from the datasets' original distributions. We find that while models trained on Tobacco-3482 and RVL-CDIP generalize well on portions of our new test datasets, they perform substantially worse on others. Moreover, models trained on Tobacco-3482 exhibit far less capacity for generalizability than those trained on RVL-CDIP.

\section{Related Work}
Measuring out-of-distribution performance is an important task that has seen increased interest in the past half-decade, especially as systems powered by machine learning make their way into more and more industries and applications. Such systems must not only perform well on in-distribution data, but must also perform well on data that is ``unseen" at training time.

Out-of-distribution performance can be measured on two different types of inputs for supervised learning tasks: first, on data that does not belong to any of the target categories of the label set; and second, on data that \emph{does} belong to the target categories but was generated or collected by a different mechanism than was used to create the original training data.

Examples of the first category include so-called \emph{out-of-scope} \cite{larson_oos} and \emph{out-of-domain} \cite{coling_ood} data in the dialog system field. Prior work has also analyzed out-of-distribution performance in image classification tasks where models trained on CIFAR datasets are tested on a wide range of data that do not belong to any of CIFAR's target categories \cite{odin}.

Prior work from the second out-of-distribution category include \cite{transformers_calibration} and \cite{hendrycks_acl_2020}, who evaluate transformer models on the NLP tasks of natural language inference, sentiment analysis, and news article classification. Other related work on \emph{out-of-domain} sentiment analysis includes \cite{cross_domain_2018} and \cite{lottery_2019}, who train sentiment analysis models on product reviews for a specific product category, then analyze model performance on reviews from a different category. In this way, the analysis of the second type of out-of-distribution phenomenon is the same as what is known as \emph{distribution shift}, where the ``training distribution differs from the test distribution" \cite{wilds_2021}. In our current paper, we focus only on this second category of out-of-distribution inference.

\begin{figure}
\includegraphics[width=\textwidth]{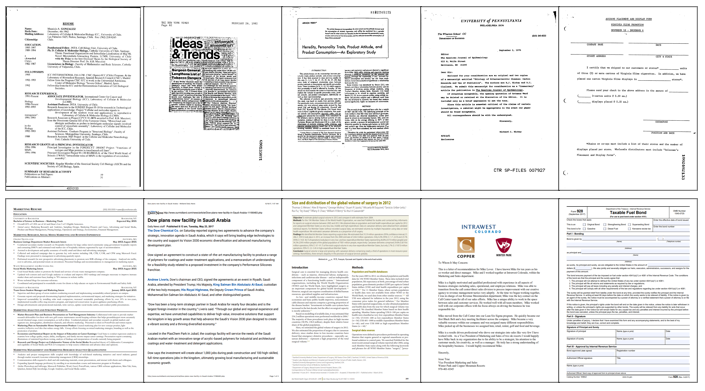}
\caption{Samples from the Tobacco-3482 corpus (top row) and our out-of-distribution evaluation corpus (bottom row). The labels for the samples are (from left to right): \emph{resume}, \emph{news}, \emph{scientific}, \emph{letter}, and \emph{form}.} \label{fig1}
\end{figure}

\section{In- and Out-of-Distribution Data}
\subsection{In-Distribution Data}
We use the Tobacco-3482 \cite{tobacco_3482} and RVL-CDIP \cite{rvl_cdip} datasets as \emph{in-distribution} training data for our text-based document classifiers. Documents from the Tobacco-3482 dataset belong to one of 10 document categories, while the RVL-CDIP dataset contains 16 different document categories. Both the Tobacco-3482 and RVL-CDIP datasets contain documents from the IIT-CDIP test collection \cite{itt}, which in turn contains documents from a corpus of publicly available documents from litigation against several tobacco-related companies \cite{ltdl} (the majority of these documents are from the 1950s through 2002). Examples from the Tobacco-3482 dataset can be seen in Figure~\ref{fig1} (top row).

\subsection{Out-of-Distribution Data}
We selected five categories from the Tobacco-3482 dataset and five categories from the RVL-CDIP dataset and gathered new out-of-distribution testing data for these categories. The list of selected categories is shown in Table~\ref{tab:ood_data}. We collected data from two different ``distributions" for the \emph{resume} data: first, we use internal company resumes; second, we used a search engine to scrape curriculum vitaes (CVs). (Curriculum vitaes can be thought of as long-form resumes, and alternatively 1-2 page resumes can be thought of as short-form CVs.) Importantly, we explicitly searched for CVs that did not contain the phrase ``curriculum vitae".

\begin{table}[]
    \centering
    \caption{Categories from Tobacco-3482 and RVL-CDIP covered by our out-of-distribution evaluation data sets.}
    \begin{tabular}{ccc}
    \toprule
        & & \textbf{Num. of OOD} \\
        \textbf{~~Tobacco-3482~~} & \textbf{~~RVL-CDIP~~} & \textbf{Test Samples} \\
        \midrule
        \emph{form} & \emph{form} & 215 \\
        \emph{letter} & \emph{letter} & 80 \\
        \emph{news} & \emph{news article} & 30 \\
        \emph{scientific} & --- & 133\\
        \emph{resume} & \emph{resume} & 267  \\
        --- & \emph{invoice} & 26 \\
        \bottomrule
    \end{tabular}
    \label{tab:ood_data}
\end{table}

Our out-of-distribution \emph{forms} data consists of tax forms, application forms, request forms, and donation forms that we scraped via web search. All of these forms are unfilled. Our new \emph{letter} data consists of cover letters, recommendation letters, and various other letters, all scraped via web search, as does the out-of-distribution \emph{news} and \emph{invoice} data.
The \emph{scientific} out-of-distribution data consist of academic research papers found via web search, the majority of which were sampled from several undergraduate research journals.
Examples of our new out-of-distribution documents can be seen in Figure~\ref{fig1} (bottom row).
Table~\ref{tab:ood_data} shows the number of out-of-distribution test samples that we collected for each category. Importantly, all out-of-distribution documents come from domains/areas other than the tobacco industry, and were created relatively recently (i.e., in years 2010-2021).

As seen in Table~\ref{tab:ood_data}, most of the chosen categories overlap between the Tobacco-3482 and RVL-CDIP datasets, with the exclusion of \emph{invoice} (which is not a category in Tobacco-3482) and \emph{scientific}. The \emph{scientific} data partially consists of academic research papers (an example of which is shown in the first row and third column of Figure~\ref{fig1}), which is why we use academic research papers in our out-of-distribution test set. While the RVL-CDIP corpus contains categories called \emph{scientific report} and \emph{scientific publication}, we could not determine which of these categories was more appropriate for academic research papers, and hence we do not include this in our evaluation of RVL-CDIP.

\section{Experiments}
\subsection{Methodology}
We conduct several experiments to evaluate the out-of-distribution generalization performance of text classifiers trained on the Tobacco-3482 and RVL-CDIP datasets.
First, we train models on both the Tobacco-3482 and RVL-CDIP datasets and evaluate in-distribution performance on the five categories specified in Table~\ref{tab:ood_data}. Unlike the RVL-CDIP corpus, the Tobacco-3482 dataset does not have a specified test set, so we randomly partitioned the dataset in to train and test sets with a 90-10 split.
Second, we evaluate the models by testing on our newly-collected data introduced in Section 3.2.

Third, we modify our out-of-distribution evaluation datasets by removing tokens from documents that are indicative of the name of the document's category. For instance, we remove all instances of the word ``form" and ``invoice" from the \emph{form} and \emph{invoice} categories, respectively. (All taboo words are listed in Table~\ref{tab:taboo_words}.) Following \cite{taboo}, we call this modified evaluation data the \emph{taboo} data. The purpose of the taboo evaluation data is to test whether the model can generalize further to data that does not contain features (in this case, tokens) that we suspect may be overrepresented in the training data.

\begin{table}[]
    \centering
    \caption{Taboo words for each test category.}
    \begin{tabular}{lc}
    \toprule
        \textbf{Category~~} & \textbf{Taboo Words}  \\
    \midrule
        \emph{form} & ``form"\\
        \emph{letter} & ``letter"\\
        \emph{resume} & ``resume", ``curriculum", ``vitae"\\
        \emph{news} & ``news", ``article"\\
        \emph{scientific} & ``article"\\
        \emph{invoice} & ``invoice"\\
    \bottomrule
    \end{tabular}
    \label{tab:taboo_words}
\end{table}

We use Google Tesseract as our optical character recognition (OCR) engine for extracting the in-distribution training and test text data from the RVL-CDIP and Tobacco-3482 document images. Since our out-of-distribution evaluation data is entirely from Microsoft Word and PDF files with text already embedded in the files, we use an extraction tool that simply reads and processes this text data without need for OCR.

Our text classifier of choice is MobileBERT \cite{mobilebert}, which is a pre-trained transformer model that was originally trained via knowledge distillation from BERT \cite{bert}. Since it is a product of knowledge distillation, MobileBERT consists of far fewer parameters than the original BERT model, and thus ought to be less prone to overparameterization. Nevertheless we also use label smoothing with a factor of 0.1. We train and test on the first 512 tokens of each document.

\subsection{Results}

Accuracy scores on the five in-distribution categories are shown in Table~\ref{tab:id_results}. We note that the Tobacco-3482 dataset appears to be easier than the RVL-CDIP dataet, as the classifier's accuracy scores on the former are higher in all four categories in which there is a category overlap.

\begin{table}[]
    \centering
    \caption{In-distribution accuracy scores.}\label{tab:id_results}
    \begin{tabular}{l|cccccc}
    \toprule
        \textbf{Train} & &  & ~~~~\textbf{Test} & \textbf{Data}~~~ & & \\
        \textbf{Dataset} & \textbf{\emph{~form~}} & \textbf{\emph{~letter~}} & \textbf{\emph{~resume~}} & \textbf{\emph{~news~}} & \textbf{\emph{~scientific~}} & \textbf{\emph{~invoice~}} \\
        \midrule
         Tobacco-3482~ & 93.2 & 89.5 & 100 & 89.5 & 85.2 & --- \\
         RVL-CDIP & 70.8 & 82.0 & 96.0 & 76.8 & --- & 81.2\\
         \bottomrule
    \end{tabular}
    
\end{table}

Table~\ref{tab:main_results} displays the accuracy scores for the text classifiers when evaluated on the out-of-distribution data. The results of the taboo out-of-distribution tests are shown in the second rows of each main row of Table~\ref{tab:main_results}. Compared to the in-distribution accuracy scores in Table~\ref{tab:id_results}, the out-of-distribution scores are typically lower. This is expected, as the out-of-distribution data ought to be harder for the classifier to recognize. For instance, the decline in performance on the Tobacco-3482 data tends to be quite severe (e.g. from 100 to 44.3 on \emph{resume} and 89.5 to 30.0 on \emph{news}). Surprisingly though, the out-of-distribution performance is better than the in-distribution performance in a few cases for the classifier trained on the RVL-CDIP dataset. With the exception of \emph{invoice}, the declines on the RVL-CDIP dataset from in- to out-of-distribution tend to be quite mild.

\begin{table}[]
    \centering
    \caption{Performance (accuracy) of document classifiers when trained on an in-distribution training dataset and evaluated on an out-of-distribution test dataset. The second row of each main row displays the results on the taboo out-of-distribution evaluation data.}
    \begin{tabular}{l|ccccccc}
    \toprule
    & & & & \textbf{Test} & \textbf{Data} & & \\
      \textbf{Train} & & & & &  & & \\
      \textbf{Dataset} &  \textbf{\emph{~form~}} & \textbf{\emph{~letter~}} & \textbf{\emph{~resume~}} & \textbf{\emph{~CVs~}} &  \textbf{\emph{~news~}} & \textbf{\emph{~scientific~}} & \textbf{\emph{~invoice~}} \\
    \midrule
     Tobacco-3482~ & 89.3 & 89.5 & 44.3 & 94.0 & 30.0 & 53.4 & --- \\
                   & 87.0 & 72.5 & 44.3 & 94.0 & 26.7 & 53.4 & --- \\
     \midrule
     RVL-CDIP & 90.7 & 83.8 & 88.0 & 100 & 73.3 & --- & 69.2 \\
              & 86.5 & 82.5 & 88.0 & 100 & 73.3 & --- & 46.2 \\
    \bottomrule
    \end{tabular}
    \label{tab:main_results}
\end{table}

In all cases, we either see a decline or no change in performance when comparing the out-of-distribution tests with the taboo tests. However, only the \emph{invoice} category from RVL-CDIP seems to be substantially impacted, dropping 23 percentage points from 69.2 to 46.2.

In general, these results indicate that the MobileBERT model trained on RVL-CDIP appears to be more robust and exhibits better generalization to out-of-distribution data than than the MobileBERT model trained on Tobacco-3482. This is likely attributed to the fact that RVL-CDIP consists of more training data. Our biggest takeaway is that models trained on RVL-CDIP do not inherently lack the capacity to generalize to documents outside of the tobacco industry or outside of documents from a fixed time period. Indeed, all of the out-of-distribution data that we collected and on which we tested comes from non-tobacco related sources and was generated relatively recently (i.e. 2010-2021). 

\section{Future Work}
Text analysis is only part of the picture in document classification, and future work will extend our analysis of out-of-distribution performance to image classification models. Future work will also evalute text- and image-based document classifiers on out-of-distribution inputs that \emph{do not} belong to any of the categories seen at training time. 

\section{Conclusion}
At the outset of this paper we were concerned with the ability of a document classification model trained on Tobacco-3482 or RVL-CDIP to generalize well to out-of-distribution documents. By training a text-based document classifier on the Tobacco-3482 and RVL-CDIP datasets and evaluating on newly-collected out-of-distribution test data, we found that while the model trained on Tobacco-3482 tends to perform poorly on out-of-distribution data, the RVL-CDIP dataset typically endows the model with the ability to generalize to this type of data.
%
%
%
%

\end{document}